\documentclass[10pt,twocolumn,letterpaper]{article}
\pdfoutput=1

\usepackage{cvpr}
\usepackage{times}
\usepackage{epsfig}
\usepackage{graphicx}
\usepackage{amsmath}
\usepackage{amssymb}
\usepackage{comment}
\usepackage{multirow}
\usepackage{caption}
\usepackage{stfloats}

\usepackage[pagebackref=true,breaklinks=true,letterpaper=true,colorlinks,bookmarks=false]{hyperref}

\newcommand{\OURS}{Scan2Mesh}

\def \path{\bp C}

\cvprfinalcopy 

\def\cvprPaperID{XXXX} 

\ifcvprfinal\fi
\begin{document}

\title{\OURS: From Unstructured Range Scans to 3D Meshes}

\author{
	Angela Dai \hspace{2cm} Matthias Nie{\ss}ner \vspace{0.1cm} \\
	Technical University of Munich \\
}

\twocolumn[{%
	\renewcommand\twocolumn[1][]{#1}%
	\maketitle
	\begin{center}
		\includegraphics[width=\linewidth]{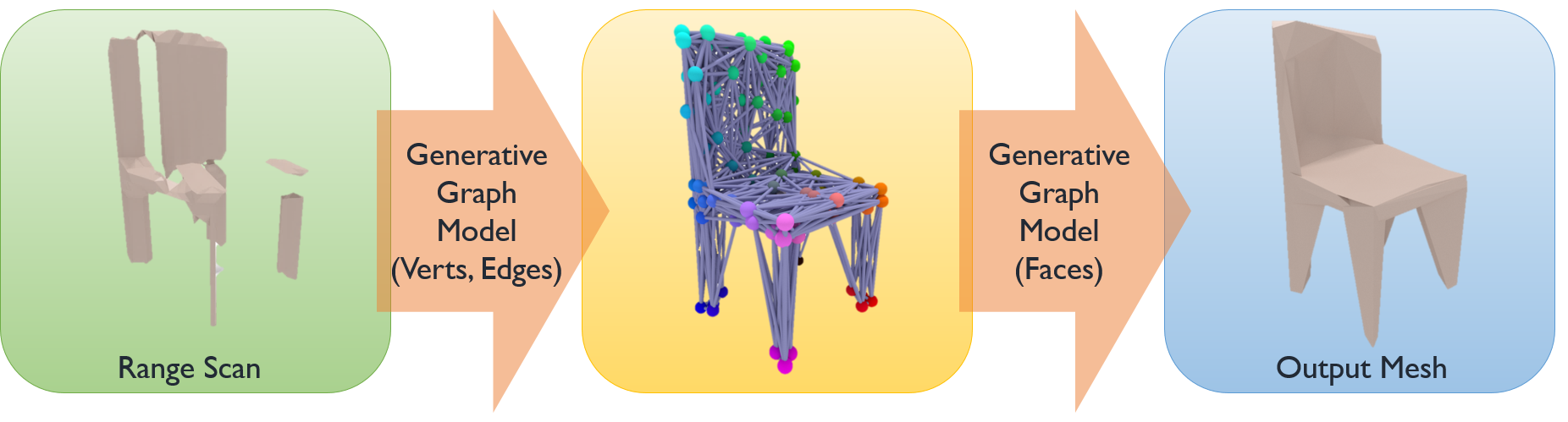}
		\captionof{figure}{3D scans of objects suffer from sensor occlusions as well as noisy, oversmoothed reconstruction quality in very dense, triangle-heavy meshes, due to both sensor noise and resolution as well as reconstruction artifacts.
			We propose a novel approach, leveraging graph neural networks, which takes a partial scan of an object and generates a complete, lightweight 3D mesh of the object.
			Our approach is the first to propose a generative deep-learning based model for directly creating a 3D mesh as an indexed face set.}
		\label{fig:teaser}
	\end{center}    
	\vspace{0.5cm}
}]

\maketitle

\begin{abstract}
We introduce \OURS{}, a novel data-driven generative approach which transforms an unstructured and potentially incomplete range scan into a structured 3D mesh representation.
The main contribution of this work is a generative neural network architecture whose input is a range scan of a 3D object and whose output is an indexed face set conditioned on the input scan.
In order to generate a 3D mesh as a set of vertices and face indices, the generative model builds on a series of proxy losses for vertices, edges, and faces. 
At each stage, we realize a one-to-one discrete mapping between the predicted and ground truth data points with a combination of convolutional- and graph neural network architectures.
This enables our algorithm to predict a compact mesh representation similar to those created through manual artist effort using 3D modeling software.
Our generated mesh results thus produce sharper, cleaner meshes with a fundamentally different structure from those generated through implicit functions, a first step in bridging the gap towards artist-created CAD models.
\end{abstract}

\section{Introduction}
\label{sec:intro}

3D meshes are one of the most popular representations used to create and design 3D surfaces, from across content creation for movies and video games to architectural and mechanical design modeling.
These mesh or CAD models are handcrafted by artists, often inspired by or mimicking real-world objects and scenes through expensive, significantly tedious manual effort.
Our aim is to develop a generative model for such 3D mesh representations; that is, a mesh model described as an indexed face set: a set of vertices as 3D positions, and a set of faces which index into the vertices to describe the 3D surface of the model.
In this way, we can begin to learn to generate 3D models similar to the handcrafted content creation process.

The nature of these 3D meshes, structured but irregular (e.g., irregular vertex locations, varying face sizes), make them very difficult to generate.
In particular, with the burgeoning direction of generative deep learning approaches for 3D model creation and completion~\cite{dai2017complete, han2017complete, song2017ssc, dai2018scancomplete}, the irregularity of mesh structures provides a significant challenge, as these approaches are largely designed for regular grids.
Thus, work in the direction of generating 3D models predominantly relies on the use of implicit functions stored in regular volumetric grids, for instance the popular truncated signed distance field representation~\cite{curless1996volumetric}. 
Here, a mesh representation can be extracted at the isosurface of the implicit function through Marching Cubes~\cite{lorensen1987marching};
however, this uniformly-sampled, unwieldy triangle soup output remains fundamentally different from 3D meshes in video games or other artist-created mesh/CAD content.

Rather than generate 3D mesh models extracted from regular volumetric grids, we instead take inspiration from 3D models that have been hand-modeled, that is, compact CAD-like mesh representations.
Thus, we propose a novel approach, \OURS{}, which constructs a generative formulation for producing a mesh as a lightweight indexed face set, and demonstrate our approach to generate complete 3D mesh models conditioned on noisy, partial range scans.
Our approach is the first, to the best of our knowledge, to leverage deep learning to fully generate an explicit 3D mesh structure.
From an input partial scan, we employ a graph neural network based approach to jointly predict mesh vertex positions as well as edge connectivity; this joint optimization enables reliable vertex generation for a final mesh structure.
From these vertex and edge predictions, interpreting them as a graph, we construct the corresponding  dual graph, with potentially valid mesh faces as dual graph vertices, from which we then predict mesh faces. 
These tightly coupled predictions of mesh vertices along with edge and face structure enable effective transformation of incomplete, noisy object scans to complete, compact 3D mesh models.
Our generated meshes are cleaner and sharper, while maintaining fundamentally different structure from those generated through implicit functions; we believe this is a first step to bridging the gap towards artist-created CAD models.

To sum up, our contributions are as follows: 
\begin{itemize}
\item A graph neural network formulation to generate meshes directly as indexed face sets.
\item Demonstration of our generative model to the task of shape completion, where we achieve significantly cleaner and more CAD-like results than state-of-the-art approaches.
\end{itemize}

\section{Related Work}
\label{sec:relatedWork}

Recent advances in machine learning, coupled with the increasing availability of 3D shape and scene databases~\cite{shapenet2015,song2017ssc,dai2017scannet}, has spurred development of deep learning approaches on 3D data.
3D ShapeNets~\cite{wu20153d} and VoxNet~\cite{maturana2015voxnet} were among the first approaches to propose 3D convolutional neural networks, both leveraging occupancy-based representations encoded in regular volumetric grids in order to perform object recognition tasks.
Various other approaches have since been developed upon 3D CNN-based architectures, targeting applications such as object classification~\cite{qi2016volumetric}, object detection~\cite{song2015deep}, 3D keypoint matching~\cite{zeng2016match}, and scan completion~\cite{dai2017complete,han2017complete,song2017ssc,dai2018scancomplete}.

Such approaches have largely been developed upon regular volumetric grid representations, a natural 3D analogue to image pixels.
Earlier 3D CNN approaches leveraged occupancy-based volumetric representations~\cite{wu20153d,maturana2015voxnet,qi2016volumetric}, simply encoding whether each voxel is occupied, empty (or optionally unknown).
Inspiration has also been taken from work in 3D scanning and reconstruction, where implicit volumetric representations, in particular truncated signed distance fields, are very popular.
Such representations encode both finer-grained information about the surface as well as the empty space, and have recently been effectively leveraged for both discriminative and generative tasks~\cite{dai2017complete,song2017ssc,dai2018scancomplete}.
For generative tasks, Liao et al.~\cite{liao2018deep} proposed a learned marching cubes mesh extraction from a volumetric grid for further output refinement.
Hierarchical strategies have also been developed to alleviate the cubic cost of such dense volumetric representations~\cite{riegler2017OctNet,wang2017cnn}, and have been shown to generate higher-resolution output grids~\cite{riegler2017octnetfusion,tatarchenko2017octree,hane2017hierarchical,han2017complete,wang2018adaptive}.
However, the 3D surfaces extracted from these regular volumetric grids maintain fundamentally different structure from that of handcrafted CAD models.

Point-based representations have recently been popularized with the introduction of the PointNet architecture~\cite{qi2017pointnet}, which demonstrated 3D classification and segmentation on a more efficient 3D representation than dense volumetric grids.
Generative approaches have also been developed upon point cloud representations~\cite{fan2017point}, but 3D point cloud outputs lack the structured surface information of meshes.

Several approaches for inferring the mesh structure of an object from an input image have recently been introduced, leveraging very strong priors on possible mesh structure in order to create the output meshes.
AtlasNet~\cite{groueix2018} learns to generate a 2D atlas embedding of the 3D mesh of an object.
Another approach is to learn to deform template meshes (e.g., an ellipsoid) to create an output 3D mesh model of an object~\cite{litany2017deformable,wang2018pixel2mesh,cmrKanazawa18}.
Such approaches generate 3D mesh surfaces as output, but are constrained to a limited set of mesh structures, whereas we aim to generate the explicit mesh structure from scratch.

In contrast to previous approaches, we take inspiration from handcrafted CAD models and develop an approach to  generate the full mesh graph structure, from vertices to edges to faces.
To this end, we leverage developments in machine learning approaches on graphs, in particular graph neural networks~\cite{scarselli2009graph}, to formulate an method to generate 3D mesh vertices, edges, and faces.

\begin{figure*}[!t] \centering
	\includegraphics[width=0.92\linewidth]{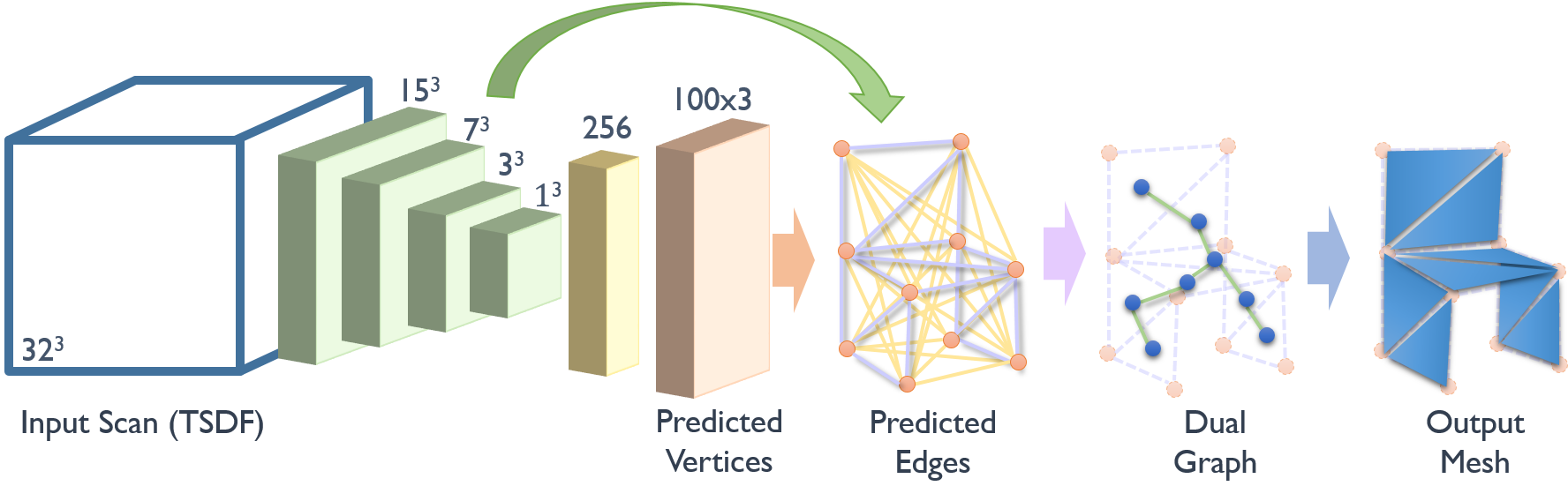}
	\vspace{-0.4cm}
	\caption{Our \OURS{} approach takes as input an partial scan of an object as a TSDF, and proposes a new graph neural network formulation to predict the mesh graph structure of vertices, edges, and faces.
		First, the input TSDF is used to jointly predict mesh vertices and edges as a graph, then this graph is transformed into its dual in order to predict the final mesh output faces (which need not contain all intermediate predicted edges). 
		We maintain losses on each of the mesh vertex, edge, and face predictions to produce a final output mesh graph structure.
		\vspace{-0.2cm}
	}
	\label{fig:architecture}
\end{figure*}

\section{Method Overview}
\label{sec:overview}

Our method generates a 3D mesh as a set of vertices (3D positions) and a set of $k$-sided faces which index into the vertices to describe the mesh surface, conditional on an input partial range scan.
Note that our approach is agnostic to the input data and representation as our focus lies in the formulation of a generative approach to explicitly generate mesh structure; in this paper, we use the task of shape completion to exemplify our approach.
For shape completion, we aim to generate a complete mesh model from an input partial scan of an object.
Here, the input scan is captured as depth image or set of depth images, which are then fused into a $32^3$ volumetric grid as a truncated signed distance field through volumetric fusion~\cite{curless1996volumetric}.
Training is performed with supervised input-target pairs, with input scans generated by virtually scanning objects from the ShapeNet dataset~\cite{shapenet2015}.

We propose a new graph neural network in order to predict the vertices, edges, and then faces of the mesh graph structure.
First, features from the input TSDF scan are computed through a series of 3D convolutions; from this feature space, we predict a set of 3D vertex locations. 
These vertex locations form the nodes of the mesh graph. 
We construct a fully connected graph on these mesh vertices, and employ graph neural network to predict which mesh edges belong to the mesh graph structure.
Note that the vertices and edges are predicted jointly in order to learn reliable vertex generation for a final mesh structure.

From the graph of intermediate predicted vertices and edges, we then construct the dual graph in order to predict the final face structure of the mesh. 
The nodes of the dual graph characterize potential faces (i.e., each node represents a potential face, which is a set of $k$ predicted edges that connect to form a valid $k$-sided face), and we employ another graph neural network to predict the final mesh faces.
We maintain losses on the vertices, edges, and faces during this mesh generation process in order to learn to generate CAD-like mesh models.

\section{\OURS{} Network Architecture}
\label{sec:arch}

Our \OURS{} network architecture is visualized in Figure~\ref{fig:architecture}.
It is composed of two main components: first, a 3D-convolutional and graph neural network architecture to jointly predict vertex locations and edge connectivity; and second, a graph neural network to predict the final mesh face structure.
For the task of shape completion, we take as input a range scan represented as a truncated signed distance field (TSDF) in a $32^3$ volumetric grid.
We represent the TSDF as a $5$-channel volumetric grid, in which the first two channels store the truncated unsigned distance field values and known/unknown space according to the camera trajectory of the scan, 
and the last three channels store the $(x,y,z)$ coordinates of the volumetric grid in the coordinate system of the mesh vertex positions, so that the TSDF volume is spatially ``aligned'' with the mesh -- in the same spirit as the CoordConv operator proposed by \cite{liu2018intriguing}.
The TSDF data generation of the partially-scanned input is detailed in Sec.~\ref{sec:datagen}.

\subsection{Joint Vertex and Edge Prediction}
The TSDF input then is used to predict a set of $n$ mesh vertex locations through a series of 3D convolutions (kernel sizes $4,3,3,3$, all but the last followed by a $1\times 1\times 1$ convolution). 
The resulting feature space, $f(\textrm{TSDF})$ is used to predict an $n\times 3$ tensor of $n$ vertex position through a series of two fully-connected layers.
We also denote the intermediary feature space after two sets of 3D convolutions as $f_2(\textrm{TSDF})$, which is used to capture spatial features of the input scan to inform the edge prediction.

We then construct a fully-connected graph with $n$ nodes corresponding to the $n$ vertex positions.
The initial node features are characterized by the $3$-dimensional vertex positions, in addition to the closest feature vector in $f_2(\textrm{TSDF})$ by looking up the vertex positions into the $f_2(\textrm{TSDF})$ grid.
We propose a graph neural network on this graph, which remains agnostic to the vertex ordering.
For a graph $\mathcal{G} = (\mathcal{V}, \mathcal{E})$ comprising vertices $v\in\mathcal{V}$ and edges $e=(v,v')\in\mathcal{E}$, messages are passed from nodes to edges, and edges to nodes as follows, similar to ~\cite{gilmer2017neural,kipf2018neural}:
\begin{equation*}
v\rightarrow e: \mathbf{h}_{i,j}' = f_e([\mathbf{h}_i, \mathbf{h}_j])
\end{equation*}
\begin{equation*}
e\rightarrow v: \mathbf{h}_i' = f_v(\sum_{\{e_{i,j}\}}\mathbf{h}_{i,j})
\end{equation*}
where $\mathbf{h}_i$ represents the feature of vertex $v_i$, $\mathbf{h}_{i,j}$ represents the feature of edge $e_{i,j}$, and $[\cdot,\cdot]$ denotes concatenation. 
Thus, an edge $e_{i,j}$ receives updates through the concatenated features of the vertices $v_i,v_j$ it is defined by, and a vertex $v_i$ receives updates through the sum of the features of the edges $e_{i,j}$ incident on $v_i$. $f_v$ and $f_e$ are MLPs operating on nodes and edges, respectively.
For full network architecture details regarding layer definitions, please see the appendix. 

\begin{figure}[!t] \centering
	\includegraphics[width=0.9\linewidth]{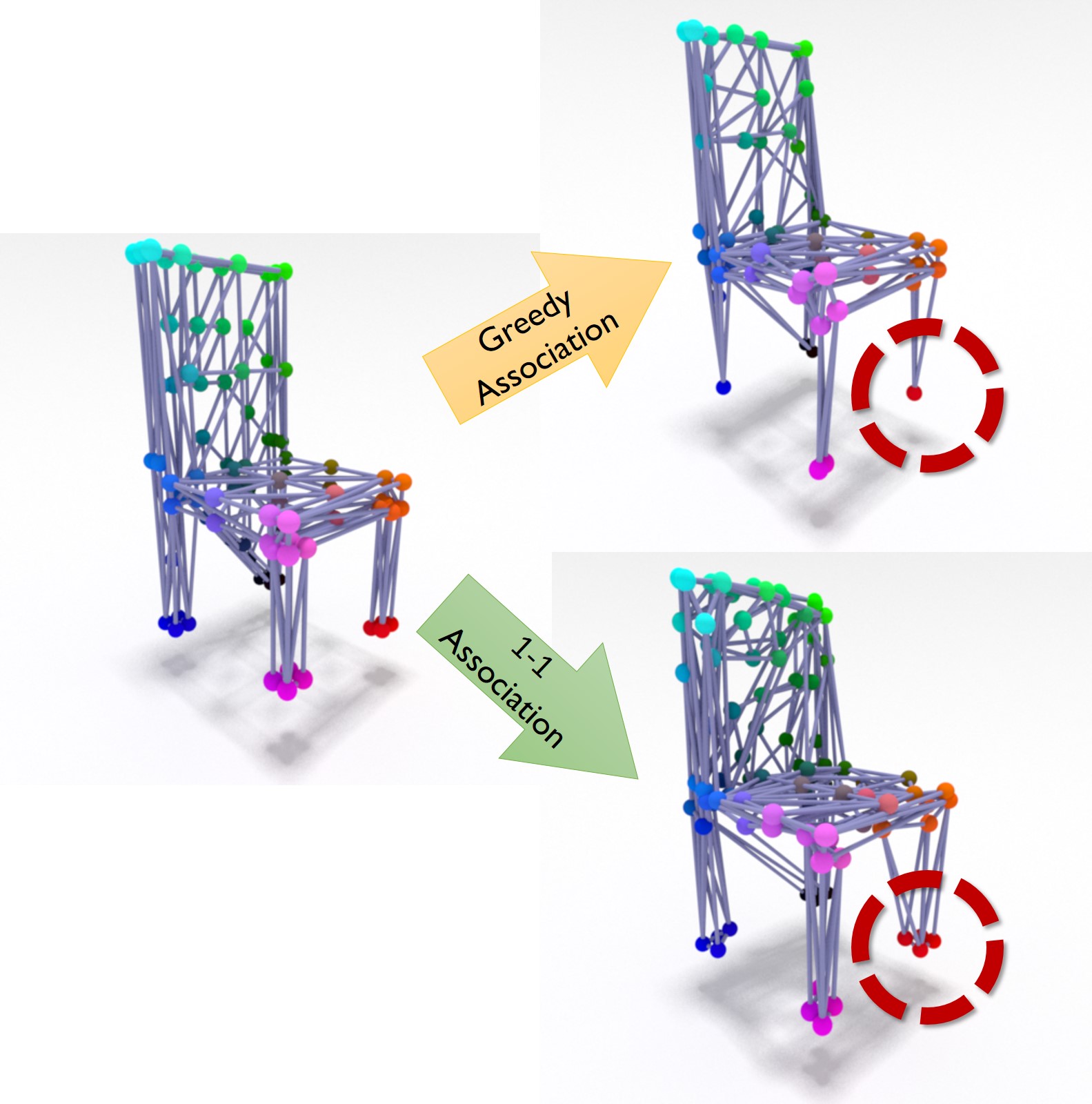}
	\vspace{-0.4cm}
	\caption{During training, we map the predicted graph with a one-to-one mapping on the vertices with the ground truth (top-left) using the Hungarian algorithm for bi-bipartite matching~\cite{kuhn1955hungarian}. This enables prediction of both large structures as well as small structures, which might collapse with a greedy association, as seen in the chair legs (top, right).
	\vspace{-0.2cm}
	}
	\label{fig:associations}
\end{figure}

The vertices are optimized for with an $\ell_1$ loss, where we first compute a one-to-one mapping between the predicted vertices and ground truth vertices using the Hungarian algorithm~\cite{kuhn1955hungarian}.
This one-to-one mapping during training is essential for predicting reliable mesh structure; a greedy approach (e.g., Chamfer loss on vertices) results in collapse of smaller structures as shown in Figure.~\ref{fig:associations}.

The output predicted vertices along with the input scan features $f_2(\textrm{TSDF})$ are then used to predict edge connectivity on the graph of the mesh with vertices as nodes.
Each node is initially associated with two features, the $3$-dimensional vertex positions and the closest feature vector in $f_2(\textrm{TSDF})$ according to the respective vertex positions.
These features are processed independently through small MLPs, then concatenated to the form vertex features which are then processed through graph message passing.
For each edge in the fully-connected graph, we predict whether it belongs to the mesh graph structure using a (two-dimensional) cross entropy loss.
The vertex positions and edges are optimized for jointly in order to reliably predict vertices belonging to a mesh structure.

\subsection{Mesh Face Prediction}
We predict the final mesh faces from these intermediate predicted vertices and edges by transforming the graph of predicted mesh vertices and edges into its dual graph.
This dual graph comprises the set of valid potential faces as the nodes of the graph, with a (dual graph) edge between two nodes if the two potential faces share an edge.
The nodes are represented by an $8$-dimensional feature vector comprising the centroid, normal, surface area, and radius of its respective potential face.
We then employ a graph neural network formulated similarly as that for the vertex and edge prediction, this time predicting which faces belong to the final mesh structure.
Note that final mesh face predictions need not contain all intermediary predicted edges.
We first train the face prediction using a cross entropy loss on the nodes of the dual graph, and then use a chamfer loss between points sampled from the predicted mesh and the target mesh in order to better encourage all structurally important faces to be predicted.

\begin{table*}[!thb]
	\begin{center}
		\small
		\begin{tabular}{| c || c | c |}
			\hline
			Method &  Mesh Distance & Mesh Normal Similarity \\  \hline
			Poisson Surface Reconstruction~\cite{kazhdan2006poisson,kazhdan2013screened} & 0.0136 & 0.60 \\ \hline
			Point Pred + Poisson~\cite{kazhdan2006poisson,kazhdan2013screened} & 0.0089 & 0.67 \\ \hline
			ShapeRecon~\cite{rock2015completing} & 0.0075 & 0.60 \\ \hline
			3D ShapeNets~\cite{wu20153d} & 0.0027 & 0.68 \\ \hline
			3D-EPN~\cite{dai2017complete} & 0.0023 & 0.76 \\ \hline
			{\bf Ours} & {\bf 0.0016} & {\bf 0.83} \\ \hline
		\end{tabular}
		\vspace{-0.3cm}
		\caption{Quantitative shape completion results for different methods on synthetic scans of ShapeNet objects.  We measure the distance between the predicted meshes and the ground truth mesh as the average point distance between points uniformly sampled from the respective meshes, as well as the normal similarity to the ground truth mesh. \emph{Point Pred + Poisson} refers to using our architecture to only predict 1024 ``vertices,'' followed by Poisson Surface Reconstruction~\cite{kazhdan2006poisson,kazhdan2013screened}.
	\vspace{-0.3cm}}
		\label{tab:quant_comparison_prev_work}		
	\end{center}
\end{table*}

\begin{table*}[!htb]
	\begin{center}
		\resizebox{\textwidth}{!}{%
			\begin{tabular}{| c || c | c || c | c | c | c | c | c | c | c | c | c | c | c | c | c | c | c |}
				\hline
				\multirow{ 2}{*}{Input} &  \multicolumn{2}{|c||}{Average}  &  \multicolumn{2}{|c|}{Chairs} &  \multicolumn{2}{|c|}{Tables} &  \multicolumn{2}{|c|}{Airplanes} & \multicolumn{2}{|c|}{Dressers} &  \multicolumn{2}{|c|}{Lamps} & \multicolumn{2}{|c|}{Boats} &  \multicolumn{2}{|c|}{Sofas}  &  \multicolumn{2}{|c|}{Cars} \\ 
				&  Dist & NSim &  Dist & NSim &  Dist & NSim &  Dist & NSim &  Dist & NSim &  Dist & NSim &  Dist & NSim &  Dist & NSim &  Dist & NSim\\  \hline
				Points & 0.0019 & 0.80 & 0.0016 & 0.79 & 0.0022 & {\bf 0.82} & {\bf 0.0008} & {\bf 0.93} & 0.0015 & 0.76  & 0.0045 & 0.72  & 0.0013 & 0.82 & 0.0021 & 0.76 & 0.0009 & 0.78\\ \hline
				TSDF & {\bf 0.0016} & {\bf 0.83} & {\bf 0.0015} & {\bf 0.82}  & {\bf 0.0021} & {\bf 0.82} & 0.0010 & {\bf 0.93} & {\bf 0.0014} & {\bf 0.79} & {\bf 0.0029} & {\bf 0.80} & {\bf 0.0011 } & {\bf 0.85 }  & {\bf 0.0016} & {\bf 0.80} & {\bf 0.0008} & {\bf 0.83}\\ \hline
			\end{tabular}
		}
	\vspace{-0.3cm}
		\caption{Evaluating the effect of different input scan representations. We compare point cloud inputs with TSDF inputs, measuring the distance between the predicted meshes and the ground truth mesh as the chamfer distance between points uniformly sampled from the respective meshes, as well as the normal similarity to the ground truth mesh. 
			The regularity of the TSDF and encoding of known and unknown space result in improved mesh prediction results.
	\vspace{-0.2cm}}
		\label{tab:quant_comparison_input_representation}		
	\end{center}
	\vspace{-0.6cm}
\end{table*}

\subsection{Training}
\label{sec:training}

To train our model, we use the training data generated from the ShapeNet dataset~\cite{shapenet2015} as described in Sec.~\ref{sec:datagen}.

We use the ADAM optimizer~\cite{kingma2014adam} with a learning rate of $0.0005$ and batch size of 8 for all training. 
We train on eight classes of the ShapeNet dataset, following the train/test split of \cite{dai2017complete}.
We additionally follow their training data augmentation, augmenting each train object by generating two virtual scanning trajectories for each object, resulting in $48,166$ train samples and $10,074$ validation samples.

We train the vertex-edge prediction for $5$ epochs ($\approx 15$ hours).
While we found it sufficient to train the joint vertex-edge prediction through finding a one-to-one mapping between the predicted vertices and ground truth mesh vertices (the edges following as vertex indices), we found that for training face prediction with cross entropy loss, the one-to-one mapping sometimes resulted in distorted target faces, and it was more reliable to train the model on dual graphs computed from the ground truth meshes.
Thus we first train the face prediction network for $1$ epoch ($\approx 6$ hours) using a cross entropy loss and ground truth dual graph data, and then train on dual graphs from predicted vertices and edges using a chamfer loss between the predicted and target meshes (for $1$ epoch, $\approx 18$ hours).

\begin{table*}[bp]
	\begin{center}
		\resizebox{\textwidth}{!}{%
			\begin{tabular}{| c || c | c || c | c | c | c | c | c | c | c | c | c | c | c | c | c | c | c |}
				\hline
				\multirow{ 2}{*}{} &  \multicolumn{2}{|c||}{Average}  &  \multicolumn{2}{|c|}{Chairs} &  \multicolumn{2}{|c|}{Tables} &  \multicolumn{2}{|c|}{Airplanes} & \multicolumn{2}{|c|}{Dressers} &  \multicolumn{2}{|c|}{Lamps} & \multicolumn{2}{|c|}{Boats} &  \multicolumn{2}{|c|}{Sofas}  &  \multicolumn{2}{|c|}{Cars} \\ 
				&  Dist & NSim &  Dist & NSim &  Dist & NSim &  Dist & NSim &  Dist & NSim &  Dist & NSim &  Dist & NSim &  Dist & NSim &  Dist & NSim\\  \hline
				Greedy & 0.0022 & 0.74 & 0.0020 & 0.73 & 0.0027 & 0.75 & 0.0012 & 0.85 & 0.0020 & 0.70 & 0.0047 & 0.70  & 0.0016 & 0.72 & 0.0021 & 0.71 & 0.0011 & 0.74\\ \hline
				1-to-1 & {\bf 0.0016} & {\bf 0.83} & {\bf 0.0015} & {\bf 0.82}  & {\bf 0.0021} & {\bf 0.82} & {\bf 0.0010} & {\bf 0.93} & {\bf 0.0014} & {\bf 0.79} & {\bf 0.0029} & {\bf 0.80} & {\bf 0.0011 } & {\bf 0.85 }  & {\bf 0.0016} & {\bf 0.80} & {\bf 0.0008} & {\bf 0.83}\\ \hline
			\end{tabular}
		}
		\vspace{-0.3cm}
		\caption{Evaluating greedy vs 1-to-1 association of predictions and ground truth during training. We measure the distance between the predicted meshes and the ground truth mesh as the chamfer distance between points uniformly sampled from the respective meshes, as well as the normal similarity to the ground truth mesh. 
			Here, a 1-to-1 discrete mapping encourages higher quality vertex, edge, and face predictions.}
		\label{tab:quant_comparison_greedy_1to1}		
	\end{center}
\end{table*}

\section{Data Generation}
\label{sec:datagen}

For training data generation, we use the ShapeNet model database \cite{shapenet2015}, and train on a subset of 8 classes (see Sec. \ref{sec:results}).
We follow the training data generation process of \cite{dai2017complete}, generating training input-target pairs by virtually scanning the ShapeNet objects along the camera trajectories given by their ShapeNet virtual scans dataset.
We use two trajectories for each object for training.
The virtually captured depth map(s) along these trajectories are then fused into a $32^3$ grid through volumetric fusion~\cite{curless1996volumetric} to obtain input TSDFs.
We use a truncation of $3$ voxels for all experiments.
An object is mapped from its world space into a $32^3$ grid by scaling the largest bounding box extent to $32 - 3*2$ (for $3$ voxels of padding on each side).

For ground truth meshes, we use triangle meshes simplified from ShapeNet models. 
In order to both reduce the complexity of the graph sizes as well as unify some of the irregularity of the ShapeNet meshes, we simplify all target meshes to $100$ vertices each using the V-HCAD library~\cite{vhcad}, which approximately maintains the convex hull of the original mesh.

\section{Results and Evaluation}
\label{sec:results}

\begin{figure*}[!thb] \centering
	\includegraphics[width=0.99\linewidth]{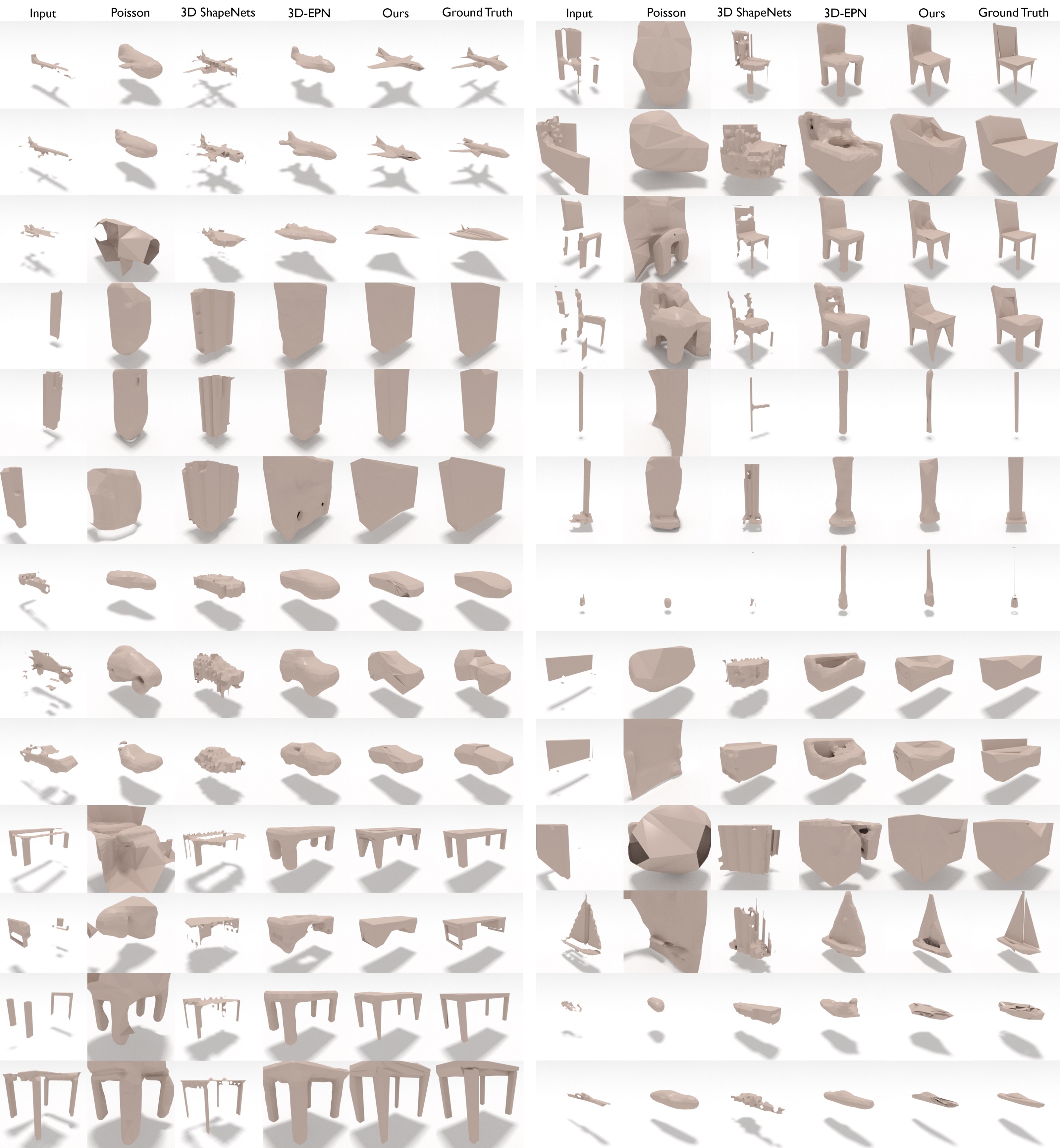}
		\vspace{-0.2cm}
	\caption{Qualitative scan completion results on virtual scans of ShapeNet~\cite{shapenet2015} objects, in comparison to Poisson Surface Reconstruction~\cite{kazhdan2006poisson,kazhdan2013screened}, as well as the volumetric generative approaches of 3D ShapeNets~\cite{wu20153d} and 3D-EPN~\cite{dai2017complete}. We show results on a variety of object classes, and produce both sharp and complete mesh structure in contrast to the volumetrically regular triangulation and noisy or oversmoothed results from approaches using implicit representations on a volumetric grid.
		\vspace{-0.4cm}
	}
	\label{fig:comparison_prevwork}
\end{figure*}

\begin{table*}[!htb]
	\begin{center}
		\resizebox{\textwidth}{!}{%
			\begin{tabular}{| c || c | c || c | c | c | c | c | c | c | c | c | c | c | c | c | c | c | c |}
				\hline
				\multirow{ 2}{*}{} &  \multicolumn{2}{|c||}{Average}  &  \multicolumn{2}{|c|}{Chairs} &  \multicolumn{2}{|c|}{Tables} &  \multicolumn{2}{|c|}{Airplanes} & \multicolumn{2}{|c|}{Dressers} &  \multicolumn{2}{|c|}{Lamps} & \multicolumn{2}{|c|}{Boats} &  \multicolumn{2}{|c|}{Sofas}  &  \multicolumn{2}{|c|}{Cars} \\ 
				&  Dist & NSim &  Dist & NSim &  Dist & NSim &  Dist & NSim &  Dist & NSim &  Dist & NSim &  Dist & NSim &  Dist & NSim &  Dist & NSim\\  \hline
				Direct(GT) & 0.0042 & 0.66 & 0.0035 & 0.62 & 0.0042 & 0.66 & 0.0078 & 0.69 & 0.0030 & 0.63 & 0.0053 & 0.67  & 0.0022 & 0.71 & 0.0058 & 0.59 & 0.0014 & 0.69 \\ \hline
				Direct(Surf) & 0.0031 & 0.69 & 0.0031 & 0.64 & 0.0028 & 0.69 & 0.0025 & 0.81 & 0.0028 & 0.62 & 0.0077 & 0.66  & 0.0016 & 0.72 & 0.0033 & 0.62 & 0.0010 & 0.73\\ \hline
				Dual Pred & {\bf 0.0016} & {\bf 0.90} & {\bf 0.0015} & {\bf 0.90}  & {\bf 0.0021} & {\bf 0.89} & {\bf 0.0010} & {\bf 0.93} & {\bf 0.0014} & {\bf 0.88} & {\bf 0.0029} & {\bf 0.86} & {\bf 0.0011 } & {\bf 0.88 }  & {\bf 0.0016} & {\bf 0.91} & {\bf 0.0008} & {\bf 0.91}\\ \hline
			\end{tabular}
		}
	\vspace{-0.3cm}
		\caption{Evaluating direct prediction of faces using a mesh graph with mesh vertices as nodes in comparison to using the dual graph with potential faces as nodes. We measure the distance between the predicted meshes and the ground truth mesh as the chamfer distance between points uniformly sampled from the respective meshes, as well as the normal similarity to the ground truth mesh. 
			The dual graph significantly reduces the large combinatorics of the possible faces, providing much more reliable mesh prediction results.
	\vspace{-0.6cm}}
		\label{tab:quant_comparison_face_pred}		
	\end{center}
\end{table*}

\begin{figure*}[tp] \centering
	\includegraphics[width=0.9\linewidth]{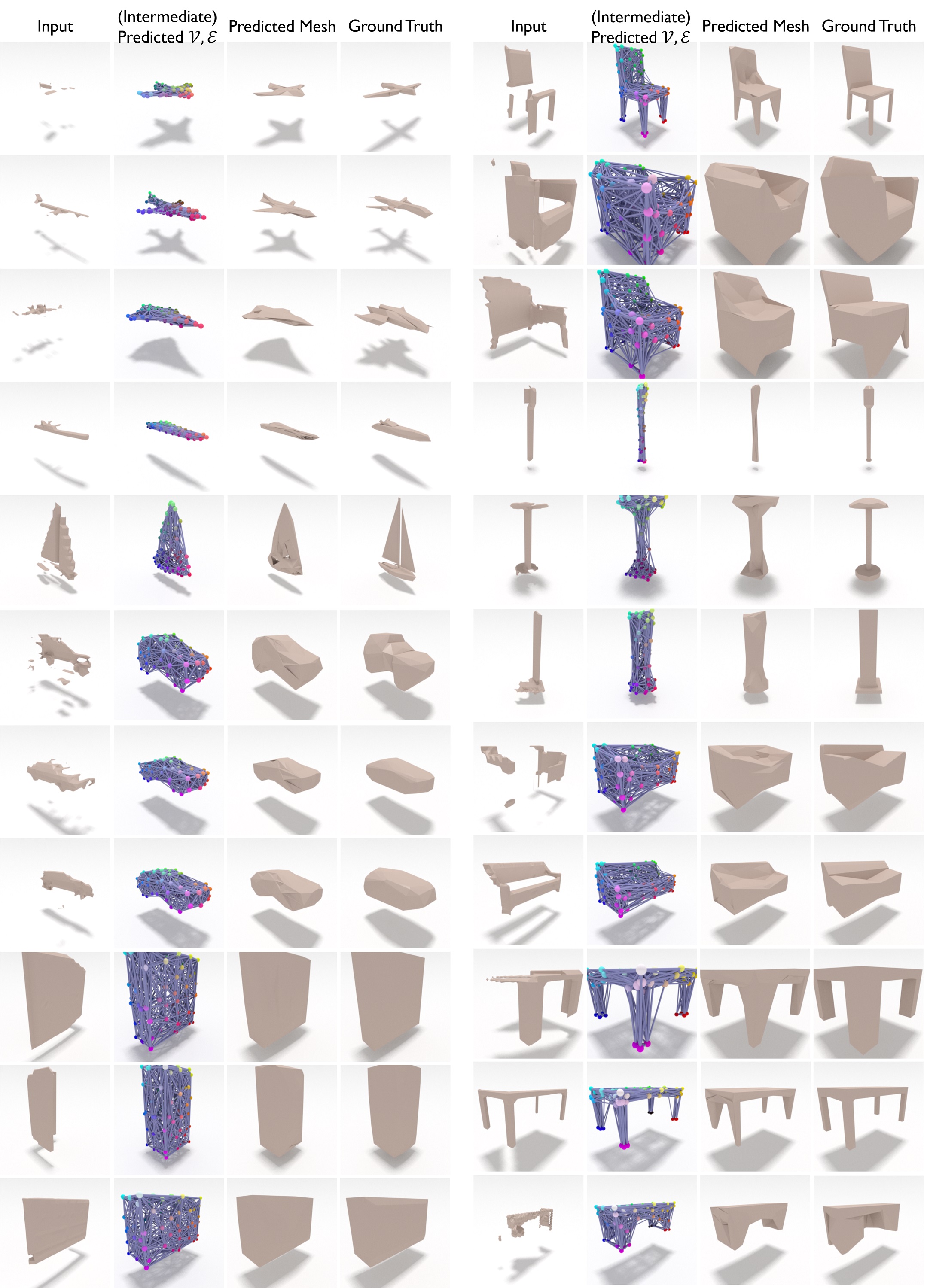}
	\caption{
		Qualitative mesh prediction results on partial scans of ShapeNet~\cite{shapenet2015} objects.
		From an input partial scan, we first predict mesh vertices and edges, which are then used to generate the final mesh face predictions.
	}
	\label{fig:gallery}
\end{figure*}

In this section, we provide an evaluation of our proposed method with a comparison to existing approaches on the task of scan completion of 3D shapes.
We evaluate on the ShapeNet~\cite{shapenet2015} dataset, using the train/test split provided by 3D-EPN~\cite{dai2017complete} comprising 8 classes: chairs, tables, sofas, dressers, lamps, boats, cars, and airplanes.
We test on the $1200$ object test set proposed by 3D-EPN of single depth image range scans ($150$ objects per class), where input scans are aligned with the ground truth complete meshes, which lie in the unit cube.
We compare our mesh results to meshes produced by state-of-the-art approaches; in the case that an approach generates an implicit function, we extract an output mesh using Matlab's {\em isosurface} function. 
To measure the mesh quality, we employ two metrics: (1) we measure the mesh completeness using a chamfer distance between uniformly sampled points from the predicted mesh and the ground truth mesh, and (2) we measure the normal deviation from the ground truth mesh to characterize mesh sharpness and cleanness. 
The normal deviation metric is computed bi-directionally: for meshes $M_a,M_b$, we sample points from each of their surfaces and compute the normal deviation $N(M_b,M_a)$ from $M_b$ to $M_a$ as the average of the cosine of the normal angle difference for the closest sampled point from $M_b$ to each that of $M_a$, and take $0.5(N(M_a,M_b) + N(M_b,M_a))$ as the global normal deviation (taking the best normal deviation from a search window of 0.03, to disambiguate small positional misalignments).

\vspace{0.2cm}\noindent
\textbf{Comparison to state of the art.}
We evaluate against state-of-the-art shape completion approaches in Table~\ref{tab:quant_comparison_prev_work} and Figure~\ref{fig:comparison_prevwork}.
Additionally, we evaluate various design choices in Tables~\ref{tab:quant_comparison_input_representation}, \ref{tab:quant_comparison_greedy_1to1}, and \ref{tab:quant_comparison_face_pred}. 
Here, we see that our approach generates sharper, cleaner meshes than previous volumetric-based approaches while producing accurate completeness in global shape structure.

\vspace{0.2cm}\noindent
\textbf{What is the impact of the input scan representation?}
We evaluate our approach using a point cloud representation of the input range scan (uniformly sampled from the range image inputs) in comparison to a TSDF in Table~\ref{tab:quant_comparison_input_representation}.
To process the point cloud input, we replace the volumetric convolutions of the encoder with a PointNet-based architecture~\cite{qi2017pointnet}.
Both representations produce good mesh results, but we find that regularity and encoding of known and unknown space in the TSDF produces better completion and mesh quality.

\vspace{0.2cm}\noindent
\textbf{Do we need a 1-to-1 mapping between prediction and target during training?}
In Table~\ref{tab:quant_comparison_greedy_1to1}, we evaluate using a greedy mapping between predicted vertices and target vertices during vertex-edge training.
Using a greedy mapping degrades the quality of vertex predictions with respect to the final mesh structure (e.g., we want a cluster of vertices at the end of a chair leg instead of one vertex), and results in worse mesh reconstruction quality (see Figure~\ref{fig:associations} for an example visualization).

\vspace{0.2cm}\noindent
\textbf{Why use the dual graph for face prediction?}
We evaluate our face prediction approach, which leverages the dual graph of the mesh vertex-edge graph, in Table~\ref{tab:quant_comparison_face_pred}. 
Here, we compare against directly predicting mesh faces using the same formulation as the joint vertex-edge prediction, where instead of predicting edges as which two vertices are connected, we predict faces as which sets of three vertices are connected, resulting in $\mathcal{O}(n^3)$ possible faces from which the mesh faces must be predicted (we refer to the appendix for more detail regarding directly predicting faces). 
Given the large combinatorics here, where the number of ground truth mesh faces is approximately $0.2\%$ of the number of total possible faces (for $n=100$), we evaluate two possibilities for training the direct face prediction: \emph{Direct(GT)} uses only the ground truth mesh faces as target faces, and \emph{Direct(Surf)} which uses all possible faces close to the ground truth mesh surface as target faces.
Both approaches nonetheless suffer from the heavy combinatorics, whereas our approach of predicting faces by using the dual graph of the mesh vertex-edge graph produces significantly better mesh structure and completeness.

 \subsection{Limitations}
 
 We propose one of the first approaches to explicitly generate a 3D mesh as an indexed face set, and believe that this is a stepping stone towards future work in constructing CAD-like 3D models akin to those currently handcrafted.
 For instance, our use of fully-connected graphs limits the size of our models; adapting the graphs and message passing to enable learning on significantly larger mesh graphs would open up generation of higher resolution or larger scale models.
 Additionally, we do not explicitly enforce mesh regularity or surface continuity (which are also not given in the ShapeNet models); adding hard constraints into the optimization to guarantee these attributes would open up many more applications for these models.

\section{Conclusion}
\label{sec:conclusion}

We presented \OURS{}, a generative model for creating 3D mesh models as indexed face sets, inspired by 3D model representations used in handcrafted 3D models.
We proposed a new graph neural network formulation to generate a mesh representation directly, and demonstrated our mesh generation on the task of shape completion, achieving cleaner and more CAD-like mesh models from noisy, partial range scans.
We believe that this opens up myriad possibilities towards bridging the gap of 3D model generation towards the quality of artist-created CAD models.

\section*{Acknowledgments}
This work was supported by the ZD.B, a Google Research Grant, a TUM Foundation Fellowship, a TUM-IAS Rudolf M{\"o}{\ss}bauer Fellowship, and the ERC Starting Grant \emph{Scan2CAD}.

{\small
\bibliographystyle{ieee}
\bibliography{main}
}

\clearpage
\newpage
\begin{appendix}
\ifcvprfinal
\section{Appendix}
\fi

\begin{figure*}[bp] \centering
	\includegraphics[width=0.8\linewidth]{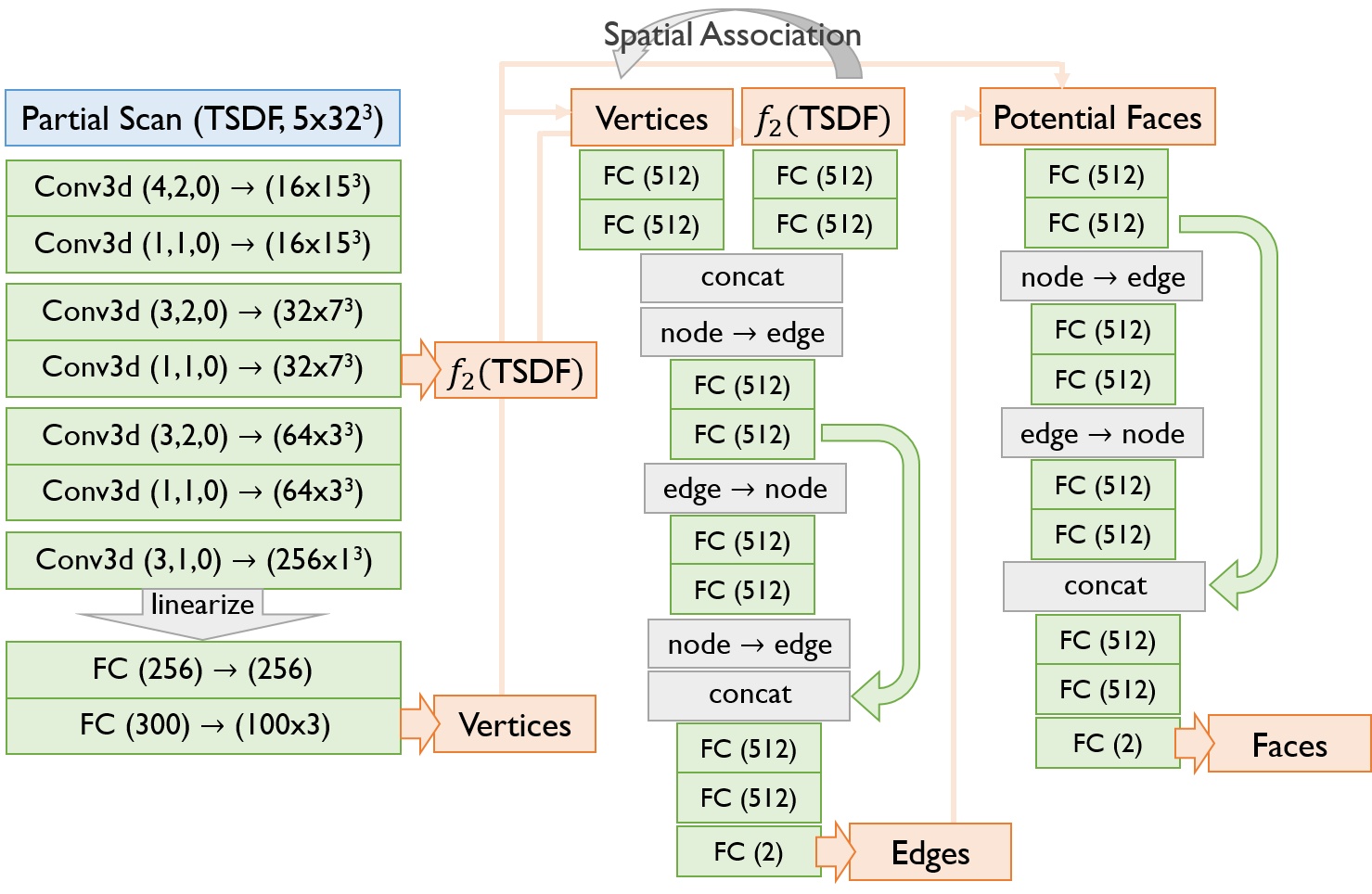}
	\vspace{-0.2cm}
	\caption{Network architecture specification for our mesh generation approach.
	}
	\label{fig:archtecture_detailed}
\end{figure*}

\ifcvprfinal
\subsection{Network Architecture Details}
\else
\section{Network Architecture Details}
\fi

Figure~\ref{fig:archtecture_detailed} shows the detailed specification of our network architecture. Convolutions are specified by (kernel size, stride, padding), and are each followed by ReLUs. For both graph neural networks, each fully connected layer (except the last) is followed by an ELU, and within each pair of fully connected layers, we use a dropout of $0.5$ during training and batch normalization following each pair.
The node-to-edge and edge-to-node message passing operations are as defined in the main paper, through concatenation and summation, respectively.

\ifcvprfinal
\subsection{Learned Feature Space}
\else
\section{Learned Feature Space}
\fi

\begin{figure*}[tp] \centering
	\includegraphics[width=0.97\linewidth]{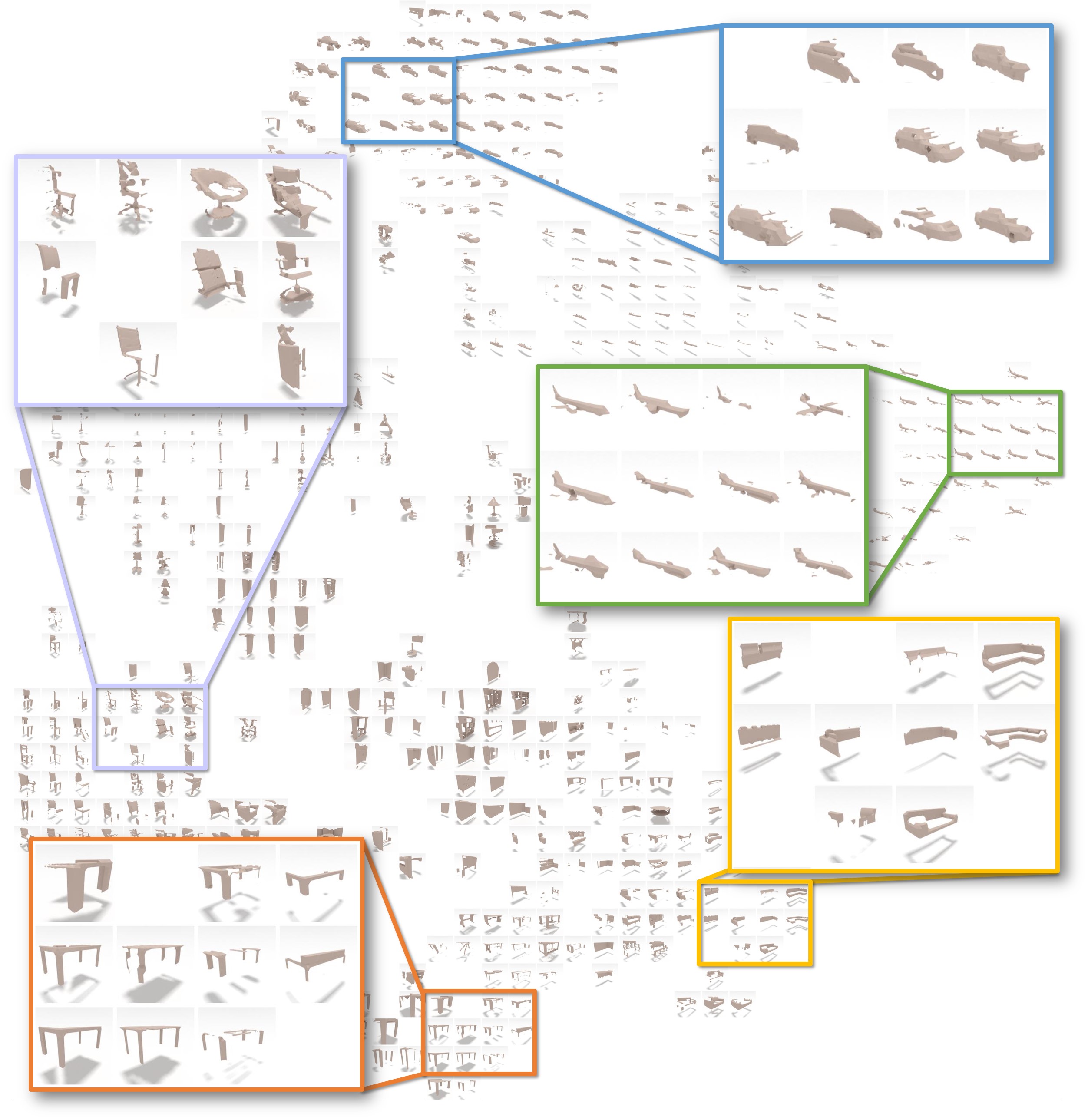}
	\vspace{-0.2cm}
	\caption{t-SNE visualization of the latent features learned from our \OURS{} model trained on shape completion. Features corresponding to input partial scans are visualized, with objects of similar geometric structure lying close together in this space.
	}
	\label{fig:tsne}
\end{figure*}

In Figure~\ref{fig:tsne}, we visualize the t-SNE of the latent vectors learned from our \OURS{} model trained for shape completion. 
We extract the latent vectors of a set of test input partial scans (the $256$-dimensional vector encoding) and use t-SNE to visualize the feature space as a 2-dimensional embedding, with images of the partial scans displayed accordingly.
Our model learns to cluster together shapes of similar geometric structure.

\begin{table}[tp]
	\begin{center}
		\resizebox{0.45\textwidth}{!}{%
			\begin{tabular}{| c || c | c || c | c |}
				\hline
				\multirow{ 2}{*}{\# Verts} & \multicolumn{2}{|c||}{Train}  &  \multicolumn{2}{|c|}{Inference} \\
				&  Time(s) & Memory(GB) &  Time(s) & Memory(GB)\\  \hline
				100 & 0.15 & 0.38 & 0.13 & 0.07 \\ \hline
				200 & 1.15 & 1.44 & 1.04 & 0.12 \\ \hline
				300 & 4.62 & 3.29 & 4.32 & 0.27 \\ \hline
				400 & 15.8 & 5.96 & 14.24 & 0.55 \\ \hline
			\end{tabular}
		}
		\caption{
			Time and memory during training and inference time for joint vertex-edge prediction as the number of predicted vertices grows. Note time measurements include a CPU hungarian algorithm computation (which currently dominates for larger number of vertices), and memory includes all allocated device memory.}
		\label{tab:numverts_time_memory}		
	\end{center}
\end{table}

\ifcvprfinal
\subsection{Shape Generation}
\else
\section{Shape Generation}
\fi

\begin{figure*}[tp] \centering
	\includegraphics[width=0.97\linewidth]{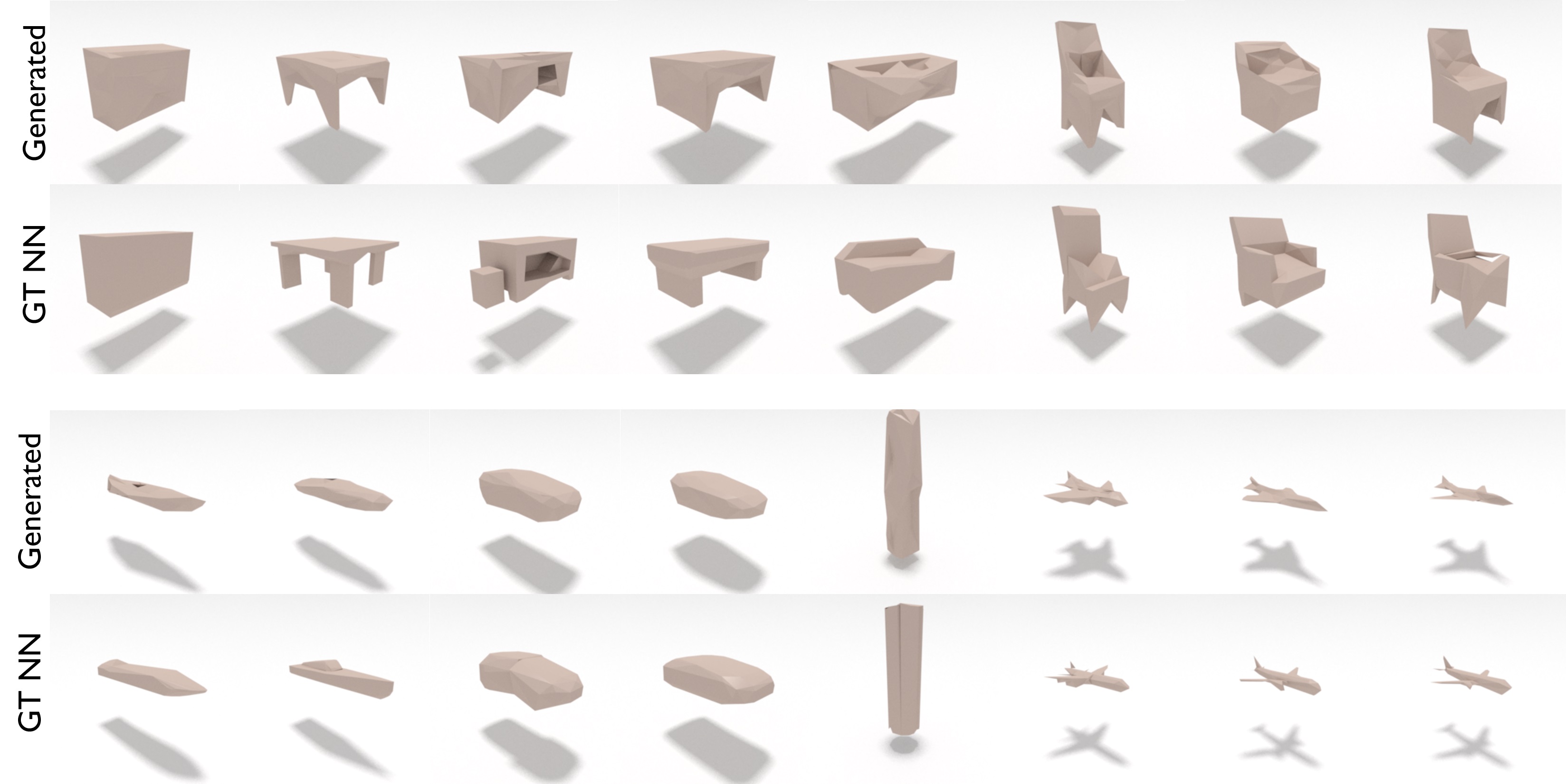}
	\caption{Our mesh generation approach applied to the task of shape generation. We show random samples from the space learned by our model, along with nearest neighbor ground truth models.
	}
	\label{fig:samples}
\end{figure*}

In the main paper, we demonstrate our mesh generation approach on the task of scan completion of shapes; we can also apply it to other tasks such as shape generation.
Here, instead of learning an encoder for TSDF scans of shapes, we train a variational autoencoder~\cite{kingma2013auto} to produce mesh vertices and edges (on the same 8-class training set of ShapeNet~\cite{shapenet2015} objects).
Figure~\ref{fig:samples} shows shapes generated by drawing random samples from a unit normal distribution, along with nearest neighbor ground truth objects.

\ifcvprfinal
\subsection{Direct Mesh Face Prediction Details}
\else
\section{Direct Mesh Face Prediction Details}
\fi

Here, we further describe the details of the approach to directly predict mesh faces from a single graph neural network, as presented in the ablation study of the main results section.
This graph network structure has the (predicted) mesh vertices as its nodes, with message passing then operating on every set of 3 nodes (assuming a triangle mesh structure).
Messages are then passed from node to face through concatenation, and from face to node through summation, similar to the node-edge message passing:
\begin{equation*}
v\rightarrow f: \mathbf{h}_{i,j,k}' = g_f([\mathbf{h}_i, \mathbf{h}_j, \mathbf{h}_k])
\end{equation*}
\begin{equation*}
f\rightarrow v: \mathbf{h}_i' = g_v(\sum_{\{f_{i,j,k}\}}\mathbf{h}_{i,j,k})
\end{equation*}
where an updated face feature is the concatenation of the node features from which it is composed, and an updated node feature is the sum of features of all faces incident on that node.
Here, even for triangle meshes, the combinatorics grows tremendously with $\mathcal{O}(n^3)$, making the optimization for face structure challenging.

\end{appendix}

\end{document}